# Artificial Intelligence as Strange Intelligence:
# Against Linear Models of Intelligence

Kendra Chilson
Department of Philosophy
University of California, Riverside
Riverside, CA  92521
USA

and

Eric Schwitzgebel
Department of Philosophy
University of California, Riverside
Riverside, CA  92521
USA

February 4, 2026

# Artificial Intelligence as Strange Intelligence:
## Against Linear Models of Intelligence

Abstract: We endorse and expand upon Susan Schneider's critique of the linear model of AI progress and introduce two novel concepts: "familiar intelligence" and "strange intelligence". AI intelligence is likely to be strange intelligence, defying familiar patterns of ability and inability, combining superhuman capacities in some domains with subhuman performance in other domains, and even within domains sometimes combining superhuman insight with surprising errors that few humans would make. We develop and defend a nonlinear model of intelligence on which "general intelligence" is not a unified capacity but instead the ability to achieve a broad range of goals in a broad range of environments, in a manner that defies nonarbitrary reduction to a single linear quantity. We conclude with implications for adversarial testing approaches to evaluating AI capacities. If AI is strange intelligence, we should expect that even the most capable systems will sometimes fail in seemingly obvious tasks. On a nonlinear model of AI intelligence, such errors on their own do not demonstrate a system's lack of outstanding general intelligence. Conversely, excellent performance on one type of task, such as an IQ test, cannot warrant assumptions of broad capacities beyond that task domain.

**Strange Intelligence and Hyperintelligence**

In "The Global Brain Argument" (forthcoming), Susan Schneider describes a possible global interconnected superintelligence, the "Global Brain". Along the way, she characterizes two categories of "hyperintelligent AI": "savant" AI and superintelligence. Superintelligent systems are assumed to exceed human intelligence in virtually all domains of interest – a concept familiar in philosophy of AI since at least Bostrom's 2014 book *Superintelligence*. Savant AI systems, in contrast, leapfrog human intelligence to achieve superhuman capacity in restricted applications while falling far below human ability in other areas. Drawing on her concept of savant AI, Schneider critiques the widespread assumption that AI progress proceeds linearly from narrow AI to human-level general intelligence to superintelligence.

In this paper, we will endorse and expand upon Schneider's critique of the linear model of AI progress and introduce two novel concepts: "familiar intelligence" and "strange intelligence". AI intelligence is likely to be strange intelligence, defying familiar patterns of ability and inability, combining superhuman capacities in some domains with subhuman performance in other domains, and even within domains sometimes combining superhuman insight with surprising errors that few humans would make. We develop and defend a nonlinear model of intelligence on which "general intelligence" is not a unified capacity but instead the ability to achieve a broad range of goals in a broad range of environments, in a manner that defies nonarbitrary reduction to a single linear quantity.

We conclude with implications for adversarial testing approaches to evaluating AI capacities. If AI is strange intelligence, we should expect that even the most capable systems will sometimes fail in seemingly obvious tasks. On a nonlinear model of AI intelligence, such errors on their own do not demonstrate a system's lack of outstanding general intelligence. Conversely, excellent performance on one type of task, such as an IQ test, cannot warrant assumptions of broad capacities beyond that task domain.

*1. The Linear Model of Intelligence.*

Discussions of AI progress often assume a *linear model of intelligence* on which all intelligent beings can, at least roughly and approximately, be ranked on a unidimensional scale from less intelligent to more intelligent. This linearity concerns the ranking of capacities, not the speed of development: Predictions of exponential growth during an "intelligence explosion" do not contradict, and indeed sometimes rely on, the assumption that intelligences can be linearly ordered. An exponential curve, for example, might place time on the x axis and intelligence on the y axis. Without a linear ordering of intelligence on the y axis, it can't be *intelligence* that grows exponentially.

In this framework, humans are typically treated as an important benchmark: Humans are the textbook example of "general intelligence". AI systems that succeed only in a narrow range of tasks are treated as subhuman "narrow AI", even if they excel at those narrow tasks. "Artificial general intelligence" (AGI) is often characterized as approximately human-level, capable of sophisticated, flexible responsiveness across a wide range of domains. Finally, "superintelligence" requires the system to exceed human intelligence in practically all domains. Before critiquing this framework, let's consider it in more detail.

*Narrow Artificial Intelligence* refers to systems that can perform a fixed set of tasks, in a narrow domain, perhaps with great proficiency. Ray Kurzweil writes:

> We are well into the era of "narrow AI," which refers to artificial intelligence that performs a useful and specific function that once required human intelligence to perform, and does so at human levels or better. Often narrow AI systems greatly exceed the speed of humans, as well as provide the ability to manage and consider thousands of variables simultaneously (2005, p. 264).

Note that Kurzweil describes narrow AI as operating "at human levels *or better*" at a specific function. Narrow AI appears below us on the linear scale despite performing some tasks better than humans because it can only do one or a small range of tasks, while humans can do a much wider range. Kurzweil characterizes narrow AI as "systems that can perform particular functions that used to require the application of human intelligence" (2005, p. 92). It seems that this emphasis on "particular functions" is what really distinguishes narrow intelligence from what is usually called "general intelligence" (see below).

> Nick Bostrom also distinguishes between narrow capacities and general intelligence:
>> Many machines and nonhuman animals already perform at superhuman levels in narrow domains. Bats interpret sonar signals better than man, calculators outperform us in arithmetic, and chess programs beat us in chess. The range of specific tasks that can be better performed by software will continue to expand. But although specialized information processing systems will have many uses, there are additional profound issues that arise only with the prospect of machine intellects that have enough general intelligence to substitute for humans across the board (2014, p. 52).

Chess-playing programs like Deep Blue are a prime example of narrow AI: They outperform humans at chess, yet they cannot apply their abilities to unrelated problems. By design, they

Chilson & Schwitzgebel            December 17, 2025            Strange Intelligence, p. 5

operate in only one domain, lacking relevant data and architectures for success in other domains. Other examples include search engines such as Google, voice assistants such as Siri, and self-driving car software. Impressive as they are, these systems cannot effectively act much beyond the narrow range of their intended application. Even if we held a chessboard clearly in the camera view of an autonomous vehicle, it could not submit a valid move, nor could the cleverest chess program find its way to city hall.

Kurzweil sees the development of AI as a process of "broadening" narrow AI until it reaches humanlike capabilities:

> Through the ongoing refinement of these methods, the development of new algorithms, and the trend toward combining multiple methods into intricate architectures, narrow AI will continue to become less narrow. That is, AI applications will have broader domains, and their performance will become more flexible. AI systems will develop multiple ways of approaching each problem, just as humans do. (2005, p. 293).

Bostrom similarly sees a shift from narrow to general intelligence: "Given that machines will *eventually* vastly exceed biology in general intelligence, but that machine cognition is *currently* vastly narrower than human cognition, one is led to wonder how quickly the usurpation will take place (2014, p. 62). Bostrom makes it clear that the intellectual development of machines can be compared with human intelligence in an approximate linear ordering:

> At present, the most advanced AI system is far below the human baseline on any reasonable metric of general intellectual ability. At some point in the future, a machine might reach approximate parity with this human baseline…. The capabilities of the system continue to grow, and at some later point the system reaches parity with the combined intellectual capability of all of humanity…. Eventually, if the system's abilities continue to grow, it attains "strong superintelligence" – a level of intelligence vastly greater than contemporary humanity's combined intellectual wherewithal (2014, p. 63).

Kurzweil and Bostrom tend to characterize general intelligence, as opposed to narrow intelligence, as human-level or better. However, we still need to clarify what "general" intelligence is and how it relates to human-level intelligence.

Artificial General Intelligence (AGI) is sometimes defined as AI that has reached human-level intelligence (see Bostrom 2014, Goertzel 2015a). A typical characterization comes from Google DeepMind's recent research statement, "Taking a responsible path to AGI," which begins "Artificial general intelligence (AGI), AI that's at least as capable as humans at most cognitive tasks, could be here within the coming years" (Dragan et al. 2025, 1). As Ben Goertzel emphasizes, however, the two concepts are distinct: "AGI is a fairly abstract notion, which is not intrinsically tied to any particular characteristics of human beings" (2015a §2). The literature usually focuses on "human-level AGI", he argues, because it is a more concrete target, which facilitates evaluation and comparison. Researchers thus tend to consider "human-level" AGI when they discuss AGI benchmarks. This constrained focus is potentially misleading in two separate ways. First, as Goertzel notes, it is "difficult to place the intelligences of all possible systems in a simple hierarchy, according to which the 'intelligence level' of an arbitrary intelligence can be compared to the 'intelligence level' of a human" (2015b §2.1). Second, humanlike patterns of intellectual capacity are likely to be only a small subset of the possible manifestations of general intelligence. We will more fully characterize these two difficulties shortly. Notice that one difficulty concerns the appropriateness, or at least methodological challenge, of applying a linear metric, while the other concerns recognizing the diversity of forms that intelligence can take.

Continuing up the hypothetical scale of intelligence, "superintelligence" refers to a form of AI that vastly exceeds human capabilities across nearly all cognitive domains. This idea is often tied to concerns about existential risk, since such a system might act in ways unpredictable to humans and could threaten human survival if its goals diverge from ours. Discussion of the "alignment problem" and "safe AI" often assumes the eventual possibility of superintelligence. The idea traces to I. J. Good, a colleague of Turing, who wrote in 1965:

> Let an ultraintelligent machine be defined as a machine that can far surpass all the intellectual activities of any man however clever. Since the design of machines is one of those intellectual activities, an ultraintelligent machine could design even better machines; there would then unquestionably be an "intelligence explosion," and the intelligence of man would be left far behind…. Thus the first ultraintelligent machine is the *last* invention that man need ever make (p. 33).

Later thinkers, such as Kurzweil (2005) and Vernor Vinge (1993), popularized the notion in terms of a "technological singularity," a point at which accelerating technological change is so rapid and drastic that human life will be profoundly transformed. Bostrom formalized the term *superintelligence* in his book of the same name, defining it as "any intellect that greatly exceeds the cognitive performance of humans in virtually all domains of interest" (2014, p. 22).

*2. Difficulties with the Linear Model.*

There are two difficulties with the linear model of AI progression from subhuman narrow intelligence to human-level AGI to general superintelligence.

First, "human-level intelligence" is a shifting, vague, and arguably unprincipled standard.

Shifting: As AI systems have achieved milestones once considered quintessentially human, the benchmarks for intelligence have been revised. Bostrom notes this phenomenon in the history of game-playing AIs: What was once regarded as a pinnacle of human intellect – mastery of chess – was reclassified as insufficient once machines achieved it (Bostrom 2005, p. 11-13). As John McCarthy reputedly quipped, "As soon as it works, no one calls it AI anymore" (Meyer 2011).

Vague: It's unclear how to weigh different collections of cognitive strengths and weaknesses in terms of "general intelligence" relative to humans. Even "Narrow AI" outperforms us in many domains, not merely through faster computation but through feats such as absorbing vast amounts of text. Imagine a taxonomy of the cognitive landscape into task types, followed by a relative weighting of those task types in assessing general intelligence, followed by a quantitative assessment of a system's performance in all of those task types. Then imagine combining these assessments into a scalar score for overall general intelligence, perhaps employing a nonlinear function to ensure that sufficient incompetence with important task types generates a low score in overall intelligence even with massive success in one task type. Any attempt to build such a model to deliver a precise scalar result would be arbitrary and artificially precise. Though it might yield intelligence level 3.56 by the parameters of Model X, there will always be a Model Y that yields a very different result and is approximately as plausible a measure of general intelligence.

Arguably unprincipled: In any case, defining AGI in terms of humans is anthropocentric. Other animals, such as apes, presumably also have general intelligence, despite being bad at spelling. Why aren't apes the standard for AGI? Furthermore, every intelligent system, humans included, has blind spots: They find some tasks easy and others virtually impossible, depending on their cognitive structure and evolutionary history. Treating the particular patterns of human cognition as the standard for "general intelligence" risks an unprincipled prioritization of what our particular species is good at and de-emphasis of what we're bad at. Language acquisition, for example, comes easily to humans, but logical reasoning about abstract propositions does not. Other well-known limitations of human cognition compared to machines include our very limited working memory; our slow, attention-consuming reading and arithmetic; our inability to multitask when tasks demand deliberation and focused thought; and the speed with which we lose cognitive skills and acquired factual knowledge.

These problems could be partly addressed if there were a genuine, scientifically measurable cognitive phenomenon of general intelligence that underlies performance across a wide variety of domains. But that leads to another, deeper difficulty with linear models of general intelligence, in addition to shifting, vague, and arguably unprincipled standards of measurement. It's not clear that "general intelligence" is a robustly real phenomenon at all. As Howard Gardner for example argues, humans appear to be predisposed only to perform a variety of more specific intellectual operations (1983/1993, p. 34). In other words, every intelligent system has selective strengths and weaknesses. Gardner contrasts this empirically informed perspective with the older view that describes humans as possessing highly flexible, all-purpose mechanisms for processing information. This outdated view, according to Gardner, fails to consider our readiness for some intellectual operations and our inability to manage others (1983/1993, p. 41). Our "general" intelligence is, in fact, selective preparedness shaped by neural and evolutionary constraints.

Human vision, for example – as attempts to model and replicate it in computers have shown – is a highly sophisticated capacity. Correctly discerning the shapes and positions of objects in a cluttered environment requires extremely complex cognitive processing and appears to manifest intelligence. But it's not at all clear that such visual capacities should be understood as belonging together in a unified construct with, say, capacities at verbal puzzle solving. And yet if we exclude visual capacities from our concept of general intelligence, we exclude a huge and important part of most people's cognition and of the types of task toward which AI has historically been directed. Similar remarks apply to capacities like five-finger grasping, speech comprehension, and responding to emotional cues. All require intelligence. But if they belong to a unified construct, it can only be an artificial construct we have invented, which can be pieced together in different ways, with different weightings – a construct that might exist in our theories and imaginations but has no robustly unified instantiation in the real world.

Might "IQ" serve as a measure of general intelligence? Current language models can achieve above-human-average performance on such tests (Wasilewski and Jablonski 2024). Is human-level AGI thereby established? Of course not. Perhaps in humans, the type of puzzle solving and pattern matching capacities that IQ tests measure correlates with some more general set of capacities (though see critiques of IQ testing in Gould 1981 and Gardner 1983/1993). However, the capacity to perform well on IQ tests is decidedly not the same as a general capacity to respond effectively and flexibly to a wide range of situations in the real world. At best, IQ-test-style verbal and visual puzzle solving and pattern matching correlate, in one species, with a one particular type of cognitive capacity. IQ test performance in non-humans is not the same as *general* intelligence.

We might think of general intelligence as the ability to use information to achieve a wide range of goals in a wide variety of environments (for related definitions see Russell and Norvig 2003/2021; Legg and Hutter 2007; Chollet 2019).  And of course even the simplest entity capable of using information to achieve goals can succeed in some environments, and no finite entity could succeed in all possible goals in all possible environments.  "General intelligence" so understood is therefore a matter of degree.  Moreover, it's a *massively multidimensional* matter of degree: There are many many possible goals and many many possible environments and no non-arbitrary way to taxonomize and weight all these goals and environments into a single linear scale or definitive threshold.

It's worth noting that humans also can achieve their goals in only a very limited range of environments.  Interstellar space, the deep sea, the Earth's crust, the middle of the sky, the center of a star – transposition to any of these places will quickly defeat almost all our plans.  We depend for our successful functioning on a very specific context.  So of course do all animals and all AI systems.  Similarly, although humans are good at a certain range of tasks, we cannot detect electrical fields in the water, dodge softballs while hovering in place, communicate with dolphins by echolocation, or calculate a hundred digits of pi in our heads.  If we put a server with a language model in the desert without a power source or if we place an autonomous vehicle in a chess tournament and then interpret their incompetence as a lack of general intelligence, we risk being as unfair to them as a dolphin would be to blame us for our poor skills in their environment. Yes, there's a perfectly reasonable sense in which chess machines and autonomous vehicles have much more limited capacities than do humans.  They are narrow in their abilities compared to us by almost any plausible metric of narrowness.  But it is anthropocentric to insist that general intelligence requires generally successful performance on the tasks and in the environments that we humans tend to favor, given that those tasks and environments are such a small subset of the possible tasks and environments an entity could face. And any attempt to escape anthropocentrism by creating an unbiased and properly weighted taxonomy of task types and environments is either hopeless or liable to generate a variety of very different but equally plausible arbitrary composites.

The linear model of general intelligence thus suffers from being, in its typical application, too changeable, vague, and anthropocentric, and even worse, it does not correspond to any robust, unified phenomenon. At best, it is an artificial composite that could equally well be constructed in a variety of different ways.

*3. Schneider's Hyperintelligence and Savant Systems.*

In "The Global Brain Argument", Schneider characterizes *hyperintelligence* as a broad category that encompasses both superintelligence and "savant systems". In doing so, Schneider adds two categories to the standard taxonomy of intelligence, neither fitting on the standard linear scale.

Regarding AGI, Schneider states that on the "Standard View", AGI requires behavioral indistinguishability from a "normal human" on ordinary cognitive tasks (Schneider forthcoming, §2). She points out that this is not the only possible meaning of "general intelligence", which can also simply contrast with narrow intelligence: General intelligence can process multiple different types of information or accomplish more varied tasks. As she puts it, many beings are general intelligences in that they "integrate material across perceptual and topical domains and exhibit cognitive flexibility, responding intelligently to goals and novel situations" (Schneider forthcoming, §2). To track this difference, Schneider refers to the "Standard View" of AGI as *strict AGI*.

Schneider correctly observes that it's a mistake to assume that strict AGI is a necessary precursor to superintelligence. Even if intelligence is assumed to progress smoothly up a linear scale from subhuman to superhuman, and thus necessarily to pass through human-level along the way, human-level AGI needn't be strict AGI as Schneider defines it. For example, it might show more intelligence in some tasks and less in others in a way that balances toward an overall level of intelligence similar to a human. If intelligence can be measured on a linear scale, for something to progress from subhuman to superhuman intelligence, it must pass through a level of intelligence overall equivalent to that of a human, though not necessarily with the same exact pattern of abilities and deficits. Compare: To reach the summit of a mountain, you must pass the three-quarter mark somewhere along the way, but not necessarily by exactly the same path as other hikers. What we reject, and what Schneider might also be rejecting, is that intelligence is nonarbitrarily linearly scalable in the manner assumed. There is no well-defined summit or three-quarters mark.

Schneider explores this phenomenon when discussing *savant systems,* including modern-day LLM chatbots such as ChatGPT, Llama 2, and Gemini. She writes:

> These chatbots go beyond the highly specific, calculator-like kind of intelligence exhibited by systems that excel at Chess and Go and are moving toward something often thought to be distinctive of biological intelligence. This distinctive feature is called "general intelligence." Further, these chatbots, or agents made of collections of chatbots, already surpass us in a variety of ways.… These are savant-like skills and would have to be "dumbed down" for these chatbots to hit the strict AGI marker, even if they were improved in a range of ways to overcome their many inadequacies. But these synthetic general intelligences are also deficient in ways normal adult humans are generally not deficient (Schneider forthcoming, §2).

In other words, these LLMs exhibit a mixture of intellectual abilities: They excel at some things, far exceeding humans, while struggling with some tasks, like counting, that typical human adults see as very basic. They are neither strict AGI nor very narrow intelligences, but neither are they superintelligences. Schneider warns that if we expect to see a day when AI acts just like us as a precursor to the day of superintelligence, we might be caught off guard.

We need a different category for these systems, which are more general than paradigmatic cases of narrow AI, but which are arguably still considerably narrower than humans – depending on how we individuate environments, domains, and tasks – and which combine superhuman strengths and subhuman deficits. They fall into the class of *savant systems,* in Schneider's terms. Savant systems exhibit a mixture of intellectual abilities, some of which fall below the standard of "strict AGI" and at least of one which exceeds that standard. A hypothetical superintelligence would not be a savant system if it has no significant intellectual deficits compared to humans.

*4. An Aside on "Savants".*

Schneider's discussion of "savant systems" appears to be a reference to "savant syndrome". In humans, this phrase refers to a person who (1.) has some form of cognitive or developmental disability, and (2.) despite this, shows extraordinary ability in at least one skill area, far above the average population. Savant syndrome is not a recognized condition according to the DSM-5, but it is a widely recorded phenomenon. One famous example is Kim Peek, on whom the movie *Rain Man* was based (Weber 2009). Musical and artistic savants are widely discussed. Another widely recognized category is the "calendrical savant", who can reel off the day of the week for any arbitrary date, over a span of decades or sometimes centuries. As these examples show, the phrase "savant system" captures the same sense of a combination of significant deficits and at least one extraordinary skill.

It's worth noting, however, why savant syndrome as a concept is seen as deeply problematic by a significant portion of the community most often affected by the label, the autistic community. Autistic people have often been stereotyped as "idiot savants" (the offensive term which "savant syndrome" has replaced), apparently due to a combination of factors. There is some evidence that savant syndrome occurs more often within the autistic population than the general population, although the conflation between autism and savant abilities is often overstated. Crucially, not every highly intelligent autistic person is a savant, because they may not demonstrate any significant cognitive deficits, at least in medically recognized categories; and not every so-called "savant" is autistic or has any recognizable condition or disorder (Treffert 2005, 2014). Joseph Straus (2014) recommends that instead of medicalizing savant status as a combination of unfortunate disability and superhuman skill, we should instead see the phenomenon as part of the normal variation in human ability. Rather than see people as savants despite deficits, we should simply recognize that particular skills and interests might enable extraordinary abilities.

Human "savants", like Schneider's savant systems, simply have a different pattern of skills and deficits than is familiar from a myopic focus on the typical human case. Both challenge the naive assumption that high intellectual capacity in one area will tend to correlate with high intellectual capacity in another in a manner that facilitates dimensional reduction to a highly explanatory single linear factor of general intelligence.

*5. Intelligence in General.*

We propose treating intelligence as the capacity to achieve goals in environments through the use of information (compare Russell and Norvig 2003/2021; Chollet 2019). The broader the range of goals and environments, the more general the intelligence. These capacities obviously vary enormously from person to person and from entity-type to entity-type. Since one can have a high capacity to achieve goals of a certain type in a certain environment type and a low capacity to achieve goals of a different type in a different environment, and since the number of possible goal types and environment types is large on any reasonable accounting, intelligence exists on a massively multidimensional spectrum. Since there is no definitive taxonomy and weighting of goals and environment types, the dimensions cannot be non-arbitrarily counted and weighed against each other. And to the extent correlations among different dimensions are weak, any attempt to reduce intelligence to a linear scale will sacrifice a huge amount of information.

Research on animal intelligence yields insights both factual and methodological. Many species exhibit striking cognitive strengths, shaped by evolutionary pressures and ecological niches. These abilities are often narrow, enabling success in a limited range of tasks in specific contexts, while the species performs poorly at a wide range of other tasks that humans might value but which are less necessary for their evolutionary success. This parallels Gardner's observations about the poor correlations among different types of intellectual skills in humans. The fact of poor correlations among intellectual skill types between, and to some extent within, species justifies methodological caution about attempting to collapse intelligence into a single linear scale and about using any small number of diagnostic tests as a measure of overall general intelligence.

François Chollet (2019) has argued that AI research lacks a unified definition of intelligence, often oscillating between two conceptions. The first emphasizes goal achievement, or success in specific tasks, while the second focuses on adaptability, or flexibility across environments. These diverge when an agent excels in one domain, such as chess, while failing in others, such as language, or when an agent fails to transfer skills from one environment to related tasks in other environments – a phenomenon known as brittleness. In human intelligence research, the distinction between success and adaptability corresponds to the difference between crystallized intelligence (accumulated knowledge and skills) and fluid intelligence (the capacity for flexible reasoning and problem-solving) (Cattell 1963; Brown 2016). Historically, AI research emphasized crystallized intelligence approaches, often relying on extensive pre-programmed knowledge bases. More recent approaches in machine learning have shifted toward fluid intelligence models, emphasizing learning from experience with minimal initial knowledge. Our definition encompasses both, especially bearing in mind that success on a task type need not be instantaneous on the first try. Crystallized and fluid intelligence are very different types of capacity that cannot be straightforwardly and non-arbitrarily weighed against each other; this is another respect in which linear models of intelligence fail.

AI systems, like nonhuman animals and neuroatypical people, can combine skills and deficits in patterns that are unfamiliar to those who have attended mostly to typical human cases. AI systems are highly unlikely to replicate every human capacity, due to limits in data and optimization, as well as a fundamentally different underlying architecture. They struggle to do many things that ordinary humans do effortlessly, such as reliably interpreting everyday visual scenes and performing feats of manual dexterity. But the reverse is also true: Humans cannot perform some feats that machines perform in a fraction of a second. If we think of intelligence as irreducibly multidimensional instead of linear – as always relativized to the immense number of possible goals and environments – we can avoid the temptation to try to reach a scalar judgment about which type of entity is actually smarter and by how much.

*6. Familiar Intelligence and Strange Intelligence.*

We might think of typical human intelligence as "familiar intelligence" – familiar to us, that is – and artificial intelligence as "strange intelligence". This terminology wears its anthropocentrism on its sleeve, rather than masking it under false objectivity. Something possesses *familiar intelligence* to the degree it thinks *like us*. It is a similarity relation. How familiar an intelligence is depends on several factors. Some are architectural: What forms does the basic cognitive processing take? What shortcuts and heuristics does it rely on? How serial or parallel is it? How fast? With what sorts of redundancy, modularity, and self-monitoring for errors? Others are learned and cultural: learned habits, particular cultural practices, acquired skills, chosen effort based on perceived costs and benefits. An intelligence is *outwardly* familiar if it acts like us in intelligence-based tasks. And it is *inwardly* familiar if it does so by the same underlying cognitive mechanisms. Note that the Turing test is a test of outward familiarity in speech which assumes that the mechanisms by which outward familiarity is achieved might be inwardly unfamiliar.

Who is the in-group, the "us", for measuring familiarity? This question should be considered deliberately and explicitly, with the understanding that the familiar is not necessarily better or, in many contexts, a worthwhile goal. To a typical member of a Western Educated Industrialized Rich Democratic (WEIRD; see Henrich 2020) society, the intelligence of a typical hunter-gatherer will be less familiar than that of one's neighbors down the street. To a neurotypical college student, the intelligence of a "calendrical savant" will be less familiar than their dormmates'. The fact that "we" – however defined – find some patterns of intelligence to some degree strange is a fact about similarity relations the acknowledgement of which permits a more conscious assessment of the possibility of bias than when intelligence is treated as an objectively scalar phenomenon privileging the skills we happen to most value. We often learn more from strangers, and with the right kind of reflection we know not to judge them by local standards.

Familiarity is also a matter of degree: The intelligence of dogs is more familiar to us (in most respects) than that of octopuses. Although we share some common features with octopuses, they evolved in a very different environment and have very dissimilar cognitive architecture as a result. It's hard for us even to understand their goals, because their existence is so different. Still, as distant as our minds are from those of octopuses, we share with octopuses the broadly familiar lifeways of embodied animals who need to navigate the natural world, find food, and mate.

AI constitutes an even stranger form of intelligence. With architectures, environments, and goals so fundamentally unlike ours, AI is the strangest intelligence we have yet to encounter. AI is not a biological organism; it was not shaped by the evolutionary pressures shared by every living being on Earth, and it does not have the same underlying needs. It is based on an inorganic substrate totally unlike all biological neurophysiology. Its goals are imposed by its makers rather than being autopoietic. Such intelligence should be expected to behave in ways radically different from familiar minds.  This raises an epistemic challenge: Understanding and measuring strange intelligence may be extremely difficult for us.  Plausibly, the stranger an intelligence is from our perspective, the easier it is for us to fail to appreciate what it's up to.  Strange intelligences rely on methods alien to our cognition.

Drawing on this appreciation of alternative forms of intelligence, we propose jettisoning the narrow AI / AGI / superintelligence paradigm.  We should recognize intelligence as a massively multidimensional phenomenon that defies reduction to a linear scale, and we should acknowledge the existence of alien forms of intelligence that operate differently enough to resist straightforward comparison and understanding.  The "narrow" vs. "general" taxonomy is also potentially misleading.  All intelligences are narrow in some ways and have some degree of generality: There is no sharp break between the narrow and the general.  Schneider's hyperintelligence is a much likelier outcome than a general superintelligence that can outperform us in virtually all intellectual tasks.  As long as our architecture differs from that of AI, we will likely continue to have some patterns of strength where we continue to outperform machines, even if those machines are extremely proficient across a wide range of the types of tasks in which WEIRD graduate students and professors have historically taken pride.

*7. The Problem with Adversarial Attacks.*

In artificial intelligence research, "classifiers" are machine learning algorithms that take in data and, through some learning process, generate a set of possible labels for those data, attempting to classify the objects represented by the data set.  Common examples include image recognition programs (which take in digital images and classify the objects in those images) and elements of natural language programming like sentiment analysis (which take in text and classify the tone or sentiment of the speaker).  One common approach to making a classifier is through the use of deep neural networks (DNNs), which use a system of weighted nodes in a network to generate a probability that a given input should be classified with a given label.

Critics of DNNs often point to adversarial examples to show the ease with which these networks can be tricked into giving the wrong answers.  Adversarial examples can be subtly changed in a way that does not affect human recognition yet causes the AI to miscategorize.  For example, Christian Szegedy et al. (2014) and Chaowei Xiao et al. (2018) describe procedures for generating images that look indistinguishable to the human eye yet lead image classifiers consistently astray.

A standard interpretation of these examples is as proof that classifiers are not meaningfully "intelligent".  The objects appear completely unambiguous to the human observer.  Since humans have no problem correctly labeling the inputs, the critic contends, the classifier algorithms must lack understanding of the target category, proving that they are merely rotely applying statistical rules.  This is sometimes used as evidence that machine learning algorithms are unfit for use or that no progress is really being made toward increasing the "intelligence" of so-called AI, such as when Goodfellow et al. claimed that "These results suggest that classifiers based on modern machine learning technique…are not learning the true underlying concepts that determine the correct output label.  Instead, these algorithms have built a Potemkin village that works well on naturally occurring data, but is exposed as a fake when one visits points in space that do not have high probability in the data distribution" (2015, p. 2).  Similarly, Gary Marcus has pointed to adversarial examples to portray deep learning as "superficial," arguing that in such cases where systems appear to have learned concepts, the imputed intelligence given to them is mere "overattribution" (Marcus 2018, p. 8).

Andrew Ilyas et al., in their 2019 paper "Adversarial Examples Are Not Bugs, They Are Features", draw a different lesson from adversarial examples.  They argue that even though the examples are indistinguishable to human eyes, the data have changed in a significant way that the AI is detecting.  They experimentally demonstrate that deep learning image recognition algorithms are consistent in the features of the data they respond to in adversarial examples.  The miscategorization is not evidence of a complete lack of understanding but rather that the algorithm is responding consistently to features invisible to human observers.  Classifiers that are being "duped" by adversarial examples are responding reasonably, but using an alien reasoning process.

This example illustrates the underlying principle of strange intelligence: DNNs were built to perform certain tasks that humans associate with familiar intelligence, such as reaching the same object classification decisions that we would make.  Although impressive results have been obtained, the underlying architecture of DNNs is radically different from a human brain, and we ought to expect them to employ different methods.  Human cognition is not the only path to achieving these results.

Therefore, it should be no surprise that (1.) image classifiers can reliably identify objects under a huge range of conditions, and that (2.) they will be stumped by some cases that are easy for us.  Despite the largely similar results, since the underlying process is radically different it should fail in different ways.  This can be analogized to convergent evolution, when a useful adaptation is developed separately among different lineages of organisms dues to similar evolutionary pressures, such as the independent evolution of flight.  The tools used and specific adaptations can vary enormously, resulting in different outcomes in different conditions.

However, this is no reason to discount the general process, especially if the systems outperform humans in a complementary range of cases.

Linear assumptions about intelligence are likely implicit in some critics' dismissal of classifiers based on their susceptibility to adversarial attacks. If intelligence were linear and one-dimensional, then a single example of an egregious mistake – a mistake a human would never make, like confusing a strawberry for a toy poodle – would be enough to show that the systems are nowhere near our level of intelligence. However, since intelligence is massively multidimensional, all these cases show on their own is that these systems have certain lacunae or blindspots. Of course, we humans also have lacunae and blind spots – just consider optical illusions. Our susceptibility to optical illusions is not used as evidence of our lack of general intelligence, however ridiculous our mistakes might seem to any entity not subject to those same illusions.

*8. Conclusion.*

With Schneider, we reject the linear model of progress in artificial intelligence. Also with Schneider, we anticipate that AI systems with a broad range of intellectual capacities will have a very different profile of strengths and weaknesses than do typical humans, without passing through a phase of being overall humanlike in their range of abilities. AI systems are and will likely continue to be "strange intelligences" that operate very differently from typical humans.

We ground these ideas in a conceptualization of intelligence as massively multidimensional, diverse in its manifestations, and incapable of nonarbitrary reduction to a single dimension. "General intelligence" is not something a system either possesses or fails to possess but rather a matter of being capable of achieving a broad enough range of goals in a broad enough range of environments – a multidimensional matter of degree and always relative to the range of goals and environments of interest.

Consequences of this nonlinear approach to intelligence include:

(1.) Tests of a single capacity in an AI system should not be used to generalize about the system's overall capacities.  AI skeptics cannot justifiably use a system's bizarre-seeming failure on a single task or small range of tasks as evidence against its "general intelligence".  Similarly, AI boosters cannot use performance on a single task or small range of tasks, such as an IQ test, as evidence of a broadly humanlike suite of intellectual capacities.

(2.) We should think more creatively and expansively about the forms intelligence can take, in AI as well as human and animal cases.  As Schneider observes, too narrow a conception of intelligence risks missing potential signs of the development of a highly intelligent system whose patterns of thinking are unfamiliar to us.  A suitably broad understanding of intelligence is useful not just from an AI safety perspective but as a proper recognition of the diverse and impressive capacities of the many different types of beings who share our world.

(3.) An appropriate appreciation of strange AI intelligence also has implications for the ethical treatment of AI as potential targets of moral concern.  On the one hand, AI systems that don't possess humanlike intelligence could nevertheless be highly intelligent in their own distinct way, in a manner that is morally significant, if intelligence or reasoning capacity is assumed to be relevant to a system's intrinsic moral standing or moral considerability.  On the other hand, if intelligence is ethically significant not for its own sake but rather because it correlates with other ethically significant properties such as sentience or membership in a moral community, the theory of strange intelligence invites us to pause.  In strange intelligences, such correlations might be absent.  An open mind to strange forms of intelligence delivers greater reason to take seriously the intelligence of entities with an unfamiliar range of capacities, while complicating the question of the ethical significance of intelligence.